\def\year{2021}\relax
\documentclass[letterpaper]{article} 
\usepackage{aaai21}  
\usepackage{times}  
\usepackage{helvet} 
\usepackage{courier}  
\usepackage[hyphens]{url}  
\usepackage{graphicx} 
\usepackage{csquotes}
\usepackage{rotating}
\usepackage{natbib}  
\usepackage{caption} 
\usepackage{subcaption}
\usepackage{array}
\title{From Static to Dynamic Prediction: Wildfire Risk Assessment Based on Multiple Environmental Factors}
\author{
    Tanqiu Jiang, Sidhant K. Bendre, Hanjia Lyu, Jiebo Luo 
    \\
}
\affiliations{
    University of Rochester\\

    tjiang17@ur.rochester.edu,
sbendre@u.Rochester.edu,
     hlyu5@ur.rochester.edu,
     jluo@cs.rochester.edu
}

\begin{document}

\maketitle

\begin{abstract}
Wildfire is one of the biggest disasters that frequently occurs on the west coast of the United States. Many efforts have been made to understand the causes of the increases in wildfire intensity and frequency in recent years. In this work, we propose static and dynamic prediction models to analyze and assess the areas with high wildfire risks in California by utilizing a multitude of environmental data including population density, Normalized Difference Vegetation Index (NDVI), Palmer Drought Severity Index (PDSI), tree mortality area, tree mortality number, and altitude. Moreover, we focus on a better understanding of the impacts of different factors so as to inform preventive actions. To validate our models and findings, we divide the land of California into 4,242 grids of 0.1 degrees $\times$ 0.1 degrees in latitude and longitude, and compute the risk of each grid based on spatial and temporal conditions. To verify the generalizability of our models, we further expand the scope of wildfire risk assessment from California to Washington without any fine tuning. By performing counterfactual analysis, we uncover the effects of several possible methods on reducing the number of high risk wildfires. Taken together, our study has the potential to estimate, monitor, and reduce the risks of wildfires across diverse areas provided that such environment data is available. 
\end{abstract}

\section{Introduction}
Wildfire has made multiple big headlines in the past few years. In the year 2018 alone, more than 1.6 million acres of land were burned among 7,948 reported incidents of wildfires. At the same time, over 100 individuals had lost their lives, and more than 24,000 buildings were damaged or destroyed, which made the 2018 wildfire season the deadliest and most destructive on record in California~\cite{calfire}. The need to build effective risk prediction and mitigation models against the imminent danger is becoming increasingly urgent. More specifically, we need a better tool to alarm the government agencies and residents when there is a high risk of wildfires in a specific area so that we can plan ahead and become better prepared.

\begin{figure}
  \includegraphics[width=\linewidth]{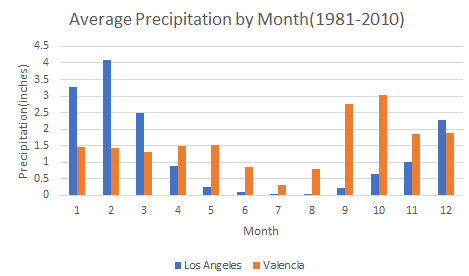}
  \caption{Average precipitation by month in Valencia, Italy and Los Angeles, U.S.}
  \label{fig1:water}
\end{figure}

The entire west coast, from San Diego to Seattle, is under Mediterranean Climate, also known as a dry-summer climate, which generally means the water precipitation is higher in winter than summer. The hot and dry summer helps create wildfire hazards in all places with Mediterranean Climate including Greece, Portugal, Italy and Spain~\cite{Molina}. Even worse, compared to the Mediterranean cities, cities in California such as Los Angeles receive a similar amount of heat from high temperature but much less rainfall during much more extended drought periods (Figure~\ref{fig1:water})\footnote{NOAA, Los Angeles/Oxnard, ``Monthly Mean Avg Temperature for Los Angeles Downtown Area, CA", https://w2.weather.gov/climate}. Consequently, California is naturally one of the most vulnerable places to wildfire because of its extremely long and dry summer. 

\section{Related works}
Wildfire of a moderate intensity is a common phenomenon in the ecosystem, as it is an essential process to create biodiversity in the forest at a large spatial scale~\cite{Safford2017NaturalRO}. However, human activities exacerbate the intensity and frequency of fire events especially at the wildland-urban interface (WUI). Back in 2007, a group of researchers investigated the relationship between human influences and fire events~\cite{human}. Using bi-variate and multiple regression methods, the research showed a highly significant relationship between population and fire frequency at the WUI. Furthermore, the research also suggested a nonlinear relationship between fire frequency and the level of human activities. The fire frequency reaches the maximum when human activities are intermediate. 

In addition to elevating the frequencies of wildfires, the vast majority of the population at the WUI often suppresses the extinguishable wildfires, while efforts to suppress large-scale fires are proven to be ineffective in most cases~\cite{North}. By excluding the benign wildfires that actually restore the ecosystem in the forests over the past century, research showed that fuels for burning (dry woods or other biomass) are being stored and making the behaviors of wildfires more unpredictable and harder to contain~\cite{Lydersen}. Apart from the accumulation of burning fuels, the densification of trees usually emerges in forests without normal wildfires, which often leads to acute competition on resources such as water and sunlight. Those that succumb to the competition dry out in the following years and become the perfect medium for fire propagation~\cite{Kolb}. Moreover, the dense forests are more vulnerable to the bark beetles~\cite{Graham}, a type of pest that feeds on the inner bark of trees, eventually killing them off one by one. The tree mortality rate has increased dramatically due to the aforementioned reasons, and the aggregation of deceased dry woods is slowly converting the forests in California into a massive pile of firelogs. 

Researchers have made attempts on solving real-world problems using different machine learning models, including predicting the wildfire risks. A group of researchers studied a small area of land in Northern California~\cite{atmos12010109}. By dividing the land into 63 1 kilometer $\times$ 1 kilometer grids, they used multiple machine learning models to predict the fire risks and then classify the studied land into five distinct risk levels. In the end, they concluded that Random Forest had the best performance in making predictions and classifying fire risks zones. In contrast, another study partitioned all the land globally into grids of 1 degree $\times$ 1 degree in latitude and longitude and used multilayer neural networks to predict the fire risks~\cite{joshi2021improving}. Their study showed that the prediction at a macro level is plausible as well, although not as accurate when analyzing at the grid-cell level.

To achieve convincing results from machine learning models, the first task is often to find appropriate predictors of wildfires. In 2018, Stephens et al.~\cite{Stephen} used Random Forest to account for the influence of topographic, weather, vegetation, and pre-fire tree mortality on the wildfire severity, and found all four variables mentioned influential. After considering all the factors mentioned above, we choose to include 1) population density, which has a significant relationship with the fire frequency, 2) Normalized Difference Vegetation Index (NDVI) measuring the densification of trees, which is vital to the forest's robustness against wildfire and bark beetles, 3) Palmer Drought Severity Index (PDSI), which measures the drought in the vicinity, 4) tree mortality area, 5) tree mortality number, and 6) altitude data as predictors to assess the risk of wildfires.


\section{Materials and Methods}
\subsection{Data Collection and Preprocessing}
In this section, we describe the data we collect and the necessary preprocessing procedures. 

\subsubsection{Analysis unit}
As mentioned in the previous section, one study~\cite{atmos12010109} made their predictions on grids of 1km $\times$ 1km (approximately 0.01 degrees $\times$ 0.01 degrees in latitude and longitude) in a small area while another study~\cite{joshi2021improving} used 1 degree $\times$ 1 degree of spatial resolution to investigate the fire risks at the global level. To maintain the variance among the data of grids and better assess the wildfire risks at the state level, the sizes of the grids we choose for California are  between those two scales. Therefore, we divide the land of California into grids of 0.1 degrees $\times$ 0.1 degrees in latitude and longitude, and use them as the analysis units. In total, there are 4,242 grids. For each grid $Grid_{i}$, we obtain the number and area of its historical wildfires, NDVI, PDSI, tree mortality and altitude.

\subsubsection{Historical wildfire incidents}
\begin{figure}[htbp]
    \centering
    \includegraphics[width=\linewidth]{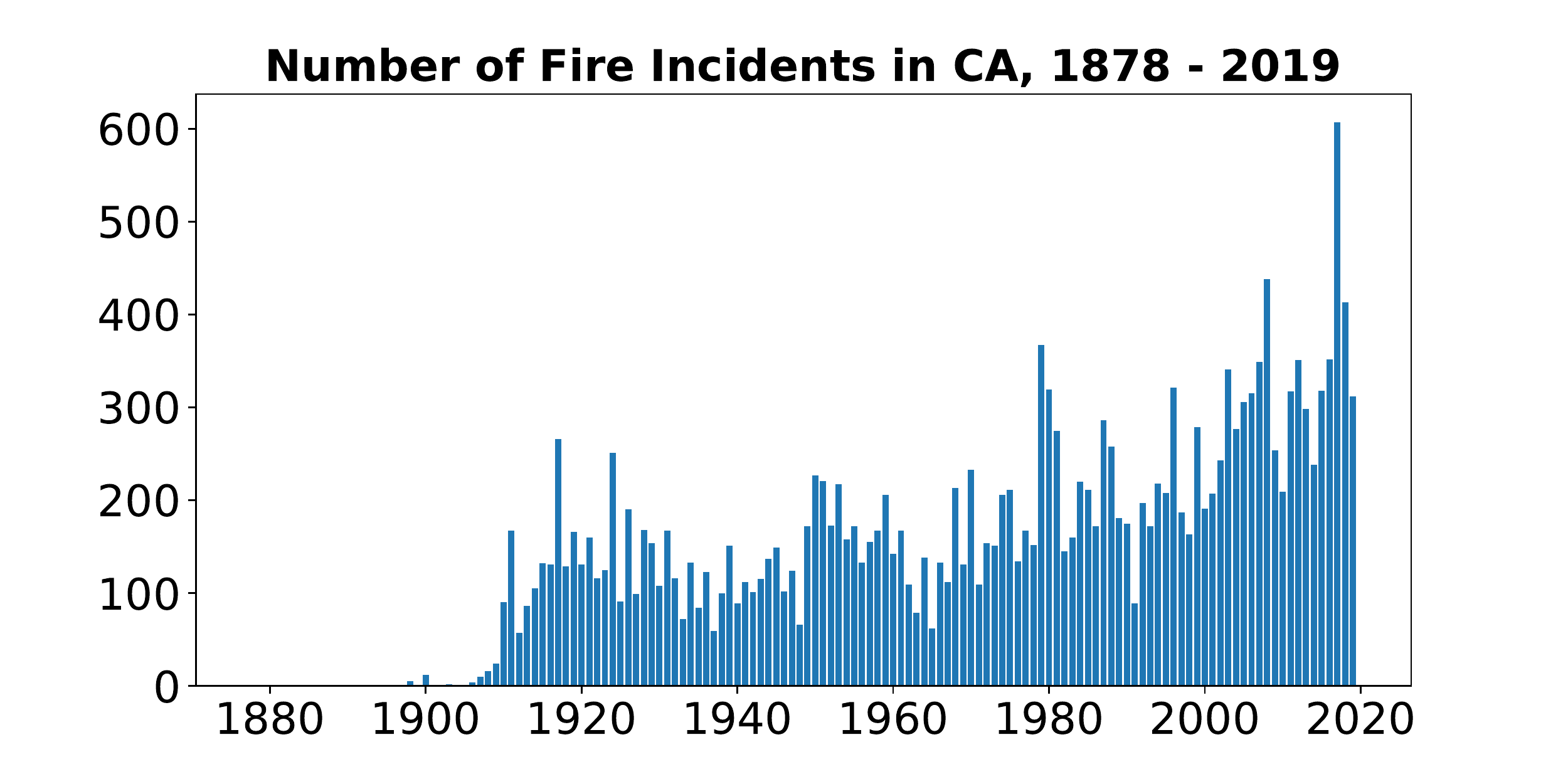}
    \caption{Number of fire incidents in CA from 1878 to 2019.}
    \label{fig:fire_incidents}
\end{figure}

We compile 20,820 recorded fire incidents from 1878 to 2019 in California by retrieving data from CA Fire~\cite{calfire}. On average, there were 146.62 fire incidents per year. Figure~\ref{fig:fire_incidents} shows the distribution of the number of fire incidents, which evinces a remarkably increasing trend. Moreover, since we attempt to estimate wildfire risk both statically and dynamically, we transform the wildfire incident information into different outcome variables, respectively.

\begin{figure}[htbp]
    \centering
    \includegraphics[width=\linewidth]{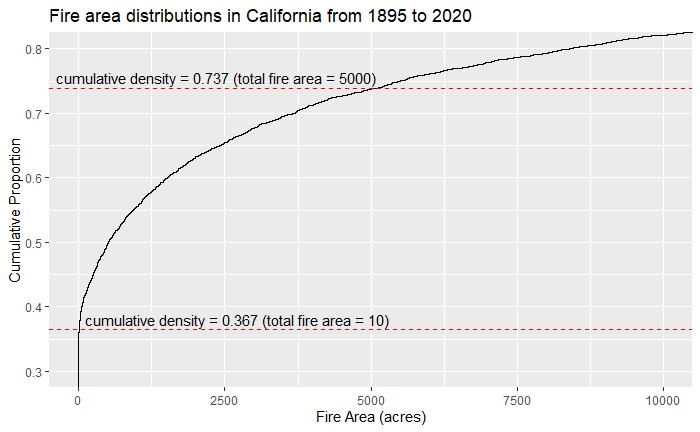}
    \caption{Cumulative distribution function.}
    \label{cdf}
\end{figure}

To build static prediction models, we sum up the areas of historical wildfires that happened in each grid, denoted by $S_{Grid_{i}}$. If $S_{Grid_{i}} <10\:acres$, the $Grid_{i}$ is labelled as a low risk zone. If $10\:acres \leq S_{Grid_{i}} <5000\:acres$, the $Grid_{i}$ is labelled as a medium risk zone. If $S_{Grid_{i}} \geq 5000\:acres$, the $Grid_{i}$ is labelled as a high risk zone. The assignment of low, medium and high risk zones is used to ensure a balanced training set. Base on the cumulative distribution plot in Figure~\ref{cdf}, 36.66\% are low risk zones, 37.03\% are medium risk zones, and the rest are high risk zones. The wildfire risk zones of CA are shown in the lower row of Figure~\ref{fig2:map}.

To build the dynamic prediction models, we assign each grid $Grid_{i,Y}$ of every year a binary label $IR_{Grid_{i,Y}}$ to indicate incoming wildfire risks. $Y$ denotes the year. As defined by the National Wildland Coordinating Group~\cite{NWCC}, any wildland fire in timber 100 acres or greater and 300 acres or greater in grasslands/rangelands is a large fire. For consistency, the threshold of large fires is set to 300 acres or larger in our study. Therefore, for $Grid_{i,Y}$, if there was at least one wildfire incident that spread more than 300 acres in $Y$, $IR_{Grid_{i,Y}}$ will be labelled 1, meaning there would be incoming wildfire risks this year, otherwise it will be labelled 0. For all 16,968 grids ranging from 2013 to 2016, 1.38\% are labelled 1 and considered incoming wildfire risk grids. The rest are labelled 0. The location of these grids are shown in the right column of Figure~\ref{fig:pre_act_ev}.


\begin{table*}[htbp]
    \centering
    \setlength{\tabcolsep}{6pt} 
    \caption{Descriptive statistics of the predictors.}
    \begin{tabular}{|c|c|c|c|c|c|c|c|c|}
    \hline
        \textbf{Predictor} & \textbf{Mean} & \textbf{Std.} & \textbf{Min.} &\textbf{25\%} &\textbf{50\%} &\textbf{75\%} &\textbf{Max.}  \\
        \hline
         Population density& 330.37&754.044&0&0.01&7.49&202.59&8445.34\\
         \hline
          NDVI& 0.43&0.23&0&0.20&0.46&0.61&0.89\\
         \hline
          PDSI (Monthly average since 1980)&-0.68 &0.41&-1.43&-1.01&-0.80&-0.26&-0.03\\
         \hline
          Tree mortality area & 2962.88&7462.58&0&0&2.90&1763.86&77342.37\\
         \hline
          Tree mortality number& 32303.40&143459&0&0&9&608&2348383\\
         \hline
          Altitude&861.06 &759.34&-89.00&200.25&717.00&1356.75&4011.00\\
         \hline
    \end{tabular}
    
    \label{tab:6_des}
\end{table*}

\begin{table*}[htbp]
\centering
\scriptsize
\caption{Sources and covered years of the predictors.}
\begin{tabular}{|c|c|c|} 
\hline
\textbf{Predictor} & \textbf{Source} & \textbf{Years covered}\\ 
\hline
Population density & SEDAC~\cite{SEDAC}&2000-2020\\ 
\hline
NDVI & MODIS~\cite{MODIS}& 2000-2015\\ 
\hline
PDSI & CDC~\cite{CDC} & 1895-2016\\ 
\hline
Tree mortality area & CAFire~\cite{calfire}& 1978-2018\\
\hline
Tree mortality number &  CAFire~\cite{calfire}& 1978-2018\\ 
\hline
Altitude & USGS~\cite{USGS} & 2000-2013\\ 
\hline
\end{tabular}
\label{tab: 6_env}
\end{table*}

\subsubsection{Predictors}
In addition to the wildfire incident data, six environmental predictors are included in this study: population density, Normalized Difference Vegetation Index (NDVI), Palmer Drought Severity Index (PDSI), tree mortality area, tree mortality number, and altitude. Table~\ref{tab:6_des} shows the descriptive statistics of the predictors. The sources and covered years of each predictor are also presented in Table~\ref{tab: 6_env}.

The population density data set is retrieved from the Socioeconomic Data and Applications Center (SEDAC)~\cite{SEDAC}. It contains the estimates of human population density (number of persons per square kilometer) based on counts consistent with national censuses and population registers, for the years 2000, 2005, 2010, 2015, and 2020 with a maximum resolution of 1 square kilometer. We map each entry of the population data into our divided 0.1 degrees $\times$ 0.1 degrees wide grids of California and then calculate the average of all the population densities in each grid as one of our predictors. 

The Normalized Difference Vegetation Index (NDVI) attribute is integrated from the Moderate Resolution Imaging Spectroradiometer (MODIS) normalized difference vegetation index (NDVI) data~\cite{MODIS}. The data set is retrieved from the NDVI data streams generated by the MODIS satellites. The data are smoothed and gap-filled to fit the American territory with the maximum resolutions. Similar to the population density attribute, our NDVI attribute is also the mean of all the NDVI data recorded on the same grid.

We use the historical PDSI data set provided by Centers for Disease Control and Prevention (CDC)~\cite{CDC} which records the Palmer Drought Severity Index by county from the year 1895 to 2016. Unlike other attributes, the drought severity index is not measured at a high resolution so we add the PDSI data to each grid based on which county it is located in. For those grids that covered the land of more than one counties, we multiply the PDSI data to the proportion of each county that the grid shares and then add the weighted PDSI of each county together as our estimated PDSI of the grid. 

The tree mortality area and tree mortality number are extracted from the same data set provided by CA Fire~\cite{calfire}. The data set records the geometric coordinates of the deceased trees, the quantity and area of the trees that were affected each year. We sum up all the tree mortality within the range of each grid in each year.

For the altitude data, we use the ASTER Global Digital Elevation Model Version 3~\cite{USGS}. The global digital elevation model provides the altitude of land areas on Earth at a spatial resolution of resolution of 1 arc second (approximately 30 meter horizontal posting at the equator). Similarly, we take the mean altitude of each 0.1 degrees by 0.1 degrees grid as our entry.

Note that the data sets from the different sources do not cover the same period of the timeline. However, some of the data are relatively steady and predictable such as altitude and population density. With some necessary modifications, we align and integrate all the data from different sources.


\subsection{Methods}
\subsubsection{Framework}
\begin{figure*}[htbp!]
    \centering
    \includegraphics[width =0.7\linewidth ]{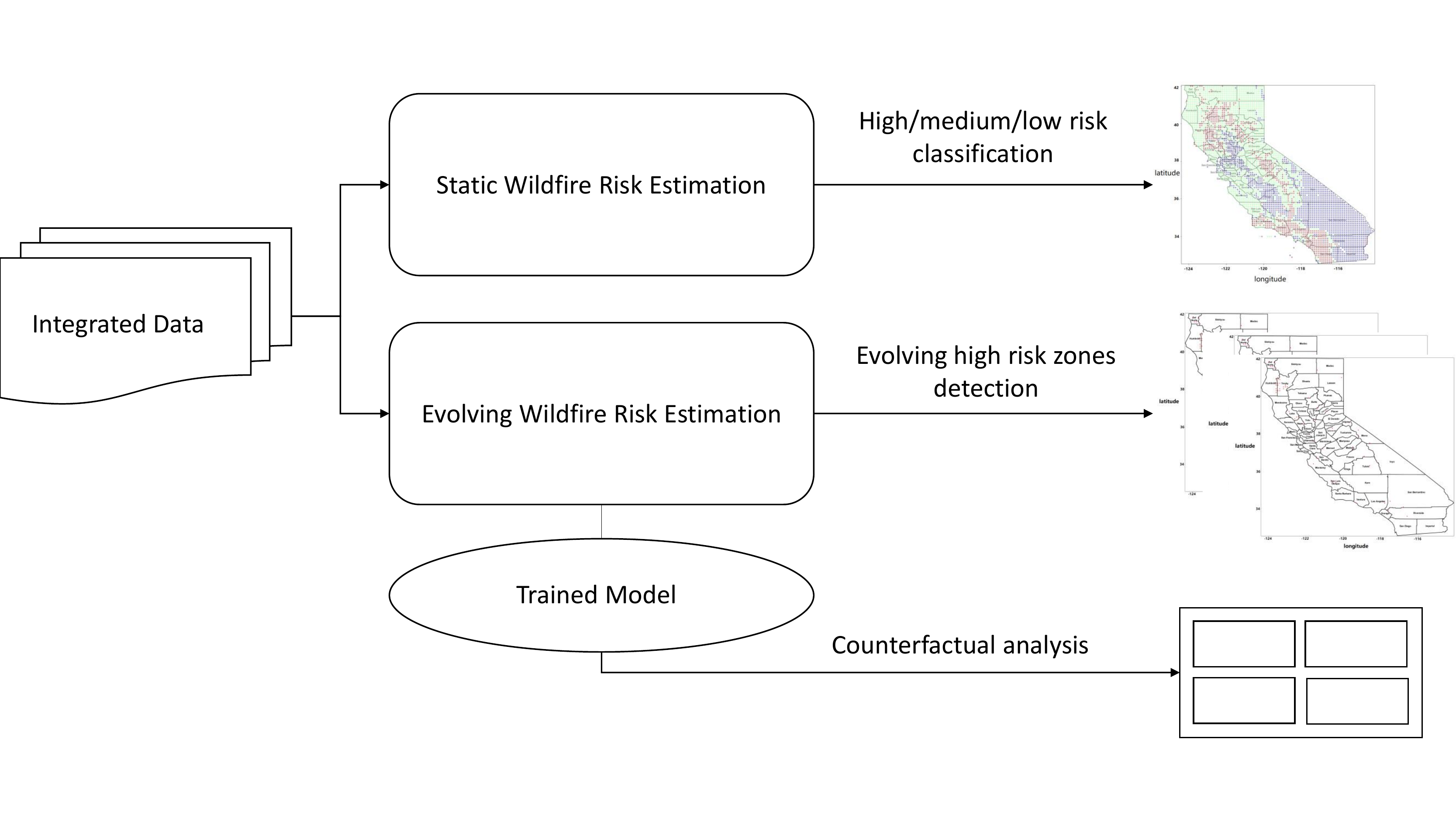}
    \caption{Diagram of static and dynamic wildfire risk assessment.}
    \label{fig:workflow}
\end{figure*}

The vast amount of data in chronological order provides us with a chance to not only estimate the risk of fire zones statically, but make predictions on the location and intensity of fires dynamically. Figure~\ref{fig:workflow} describes the workflow of this study. In a related work in 2019~\cite{JAAFARI2019198}, a research group used historical data of a small area on the north-east of Iran to plot a fire probability map. Inspired by this idea, we make static wildfire risk predictions by training three-class classification models which classify the grids of 0.1 degrees $\times$ 0.1 degrees in latitude and longitude into high, medium and low risk zones (as we define in the previous section) based on all the historical data. 

Although classifying zones into different risk levels based on historical data is a good concept to prepare us for the fire risks in the future, wildfire incidents happen in different locations every year with large variations. In other words, the risk at a specific location still varies greatly every year because of the changes in many aspects such as local weather and tree mortality. Therefore, a better tactic to prepare for the wildfires is to make predictions with respect to the specific conditions of each year. Thus, for each grid of every year, we make dynamic wildfire risk prediction by training binary classification models which predict whether there will be at least one wildfire incident that spreads more than 500 acres based on all the data before the year of prediction. 

Finally, to uncover the effect of the predictors on wildfires, we use the trained models of the dynamic wildfire risk prediction to conduct counterfactual analysis.


\subsubsection{Class imbalance}
Even though wildfire is a substantial issue on the west coast, more than 90\% of the land stays safe from fire every year. Therefore, there is an acute imbalance of classes~\cite{japkowicz2002class} when trying to make predictions in our study. As aforementioned, in the dynamic wildfire risk prediction problem, 1.38\% are positive, and the rest are negative. 
To tackle the problem, we choose to use Synthetic Minority Over-sampling Technique (SMOTE)~\cite{Chawla_2002} to balance our data while preserving as much information as possible. SMOTE over-samples the under-represented class by creating data points that are within random distances between one data point in the under-represented class and its K nearest neighbours in the same class~\cite{Chawla_2002}. In the end, the ratio of the two classes is $1:1$.

\subsubsection{Model setting}
Using a five fold cross-validation, we employ Multi-layer Neural Network (MLNN), Logistic Regression (LR), Support Vector Machine (SVM) and Random Forest (RF) to make predictions. The multi-layer Neural Network uses 6 input layer nodes (i.e., ``NDVI", ``PDSI", ``altitude", ``Population Density", ``Tree Mortality Number", ``Tree Mortality Area"), 36 hidden layer nodes, and 3 output nodes (i.e., ``high risk", ``medium risk", and ``low risk"). The learning rate is 0.01. The logistic regression model uses the ``liblinear" solver and ``L2" penalty. The SVM classifier utilizes the radial basis function kernel and ``lbfgs" solver. The Random Forest model uses ``gini" as the criterion with ``$n\_estimators = 5$". 


\section{Results}
\subsection{Static wildfire risk prediction}
\begin{figure}[htbp!]
\vspace{-2cm}
\centering
  \includegraphics[width=\linewidth]{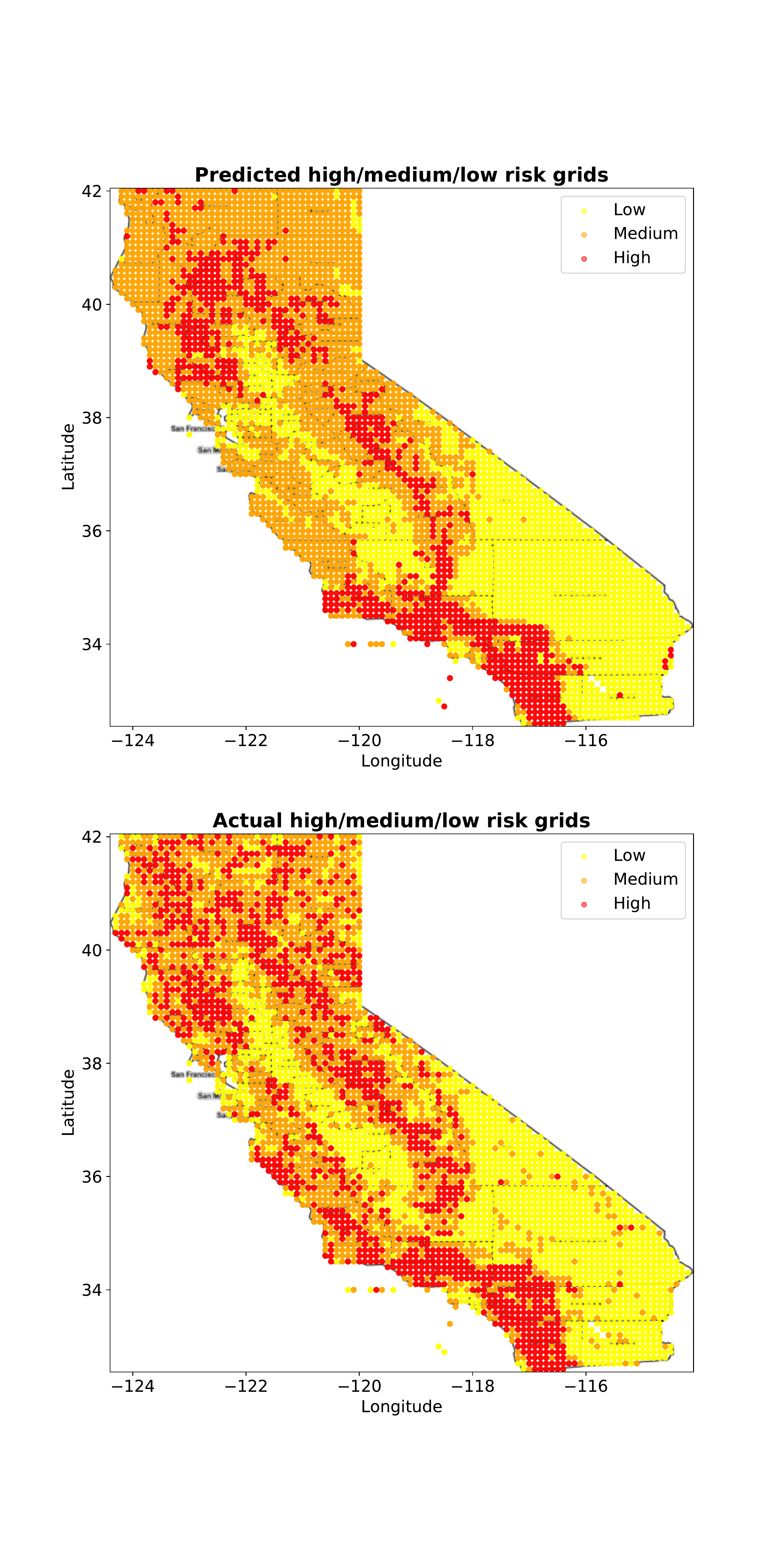}
  \caption{Predicted (upper) and actual (lower) static wildfire risk grids (red $=$ high risk, orange $=$ medium risk, yellow $=$ low risk). Qualitatively speaking, our best prediction model has high precision while recall can be improved. }
  \label{fig2:map}
\end{figure}

The classification performance is summarized in Table~\ref{tab: static_result}. MLNN has the best overall performance with 64.5\% accuracy. In general, all models have successfully classified the grids of California into low/medium/high risk zones. The predicted wildfire risk zones are shown in the upper row of Figure~\ref{fig2:map}.


\begin{table*}[htbp!]
\centering
\caption{Prediction performance.}
\begin{tabular}{|c|c|c|c|c|} \hline
Metric&MLNN&LR&SVM&RF\\ \hline
\texttt{Accuracy} & \textbf{0.645 $\pm$ 0.011}  & 0.620 $\pm$ 0.014 & 0.595 $\pm$ 0.008 & 0.617 $\pm$ 0.016\\ \hline
\texttt{Macro Precision}  & 0.621 $\pm$ 0.019  & 0.581 $\pm$ 0.024 & \textbf{0.667 $\pm$ 0.036} & 0.665 $\pm$ 0.032\\ \hline
\texttt{Macro Recall}  & \textbf{0.619 $\pm$ 0.019}  & 0.559 $\pm$ 0.024 & 0.582 $\pm$ 0.021 & 0.586 $\pm$ 0.011 \\ \hline
\texttt{Macro F1 Score}  & \textbf{0.650 $\pm$ 0.025}  & 0.548 $\pm$ 0.017  & 0.576 $\pm$ 0.013 & 0.589 $\pm$ 0.021\\ \hline

\end{tabular}
\label{tab: static_result}
\end{table*}

\subsection{Dynamic wildfire risk prediction}

\begin{figure*}[htbp!]
    \centering
    \includegraphics[width =0.78 \linewidth]{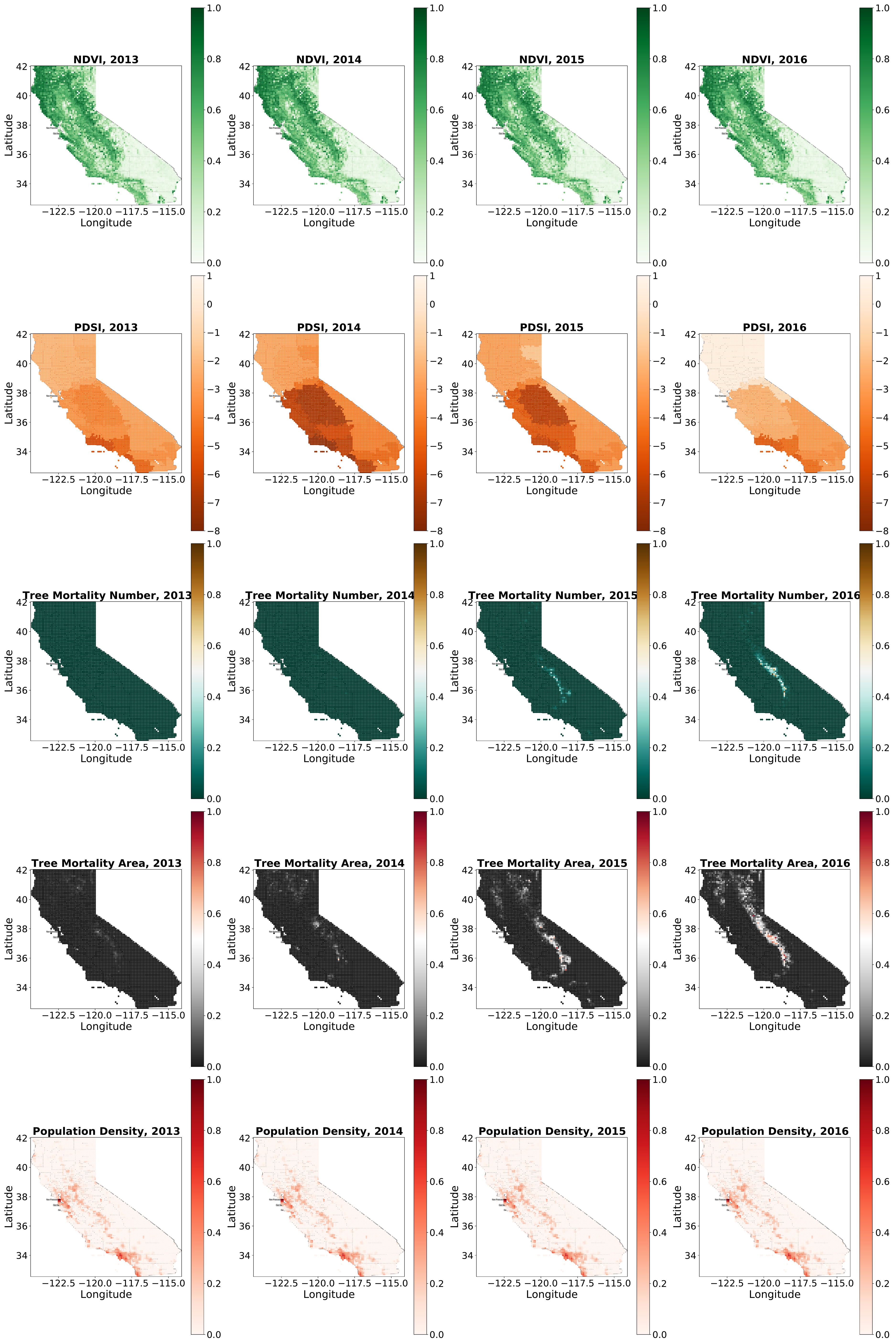}
    \caption{NDVI, PDSI, tree mortality number, tree mortality area, and population density of 2013 - 2016.}
    \label{fig:env_4_year}
\end{figure*}

Figure~\ref{fig:env_4_year} shows the changes of environmental predictors of 2013 - 2016. Since MLNN has outperformed other classifiers in static wildfire risk prediction, we use MLNN to make dynamic wildfire risk prediction. The data is split into a training set (80\%) and a testing set (20\%). 

\begin{figure}
     \centering
     \begin{subfigure}[b]{0.3\textwidth}
         \centering
         \includegraphics[width=\textwidth]{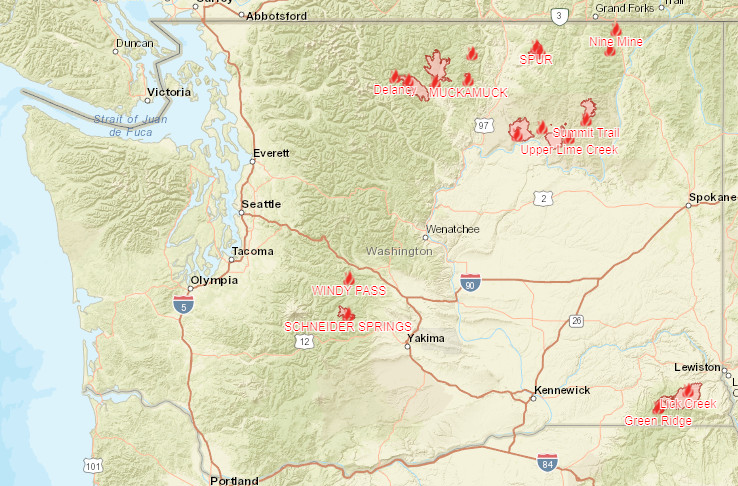}
         \caption{Actual Wildfire zone. Red indicates the places of actual wildfires.}
         \label{fig:a}
     \end{subfigure}
     \hfill
     \begin{subfigure}[b]{0.3\textwidth}
         \centering
         \includegraphics[width=\textwidth]{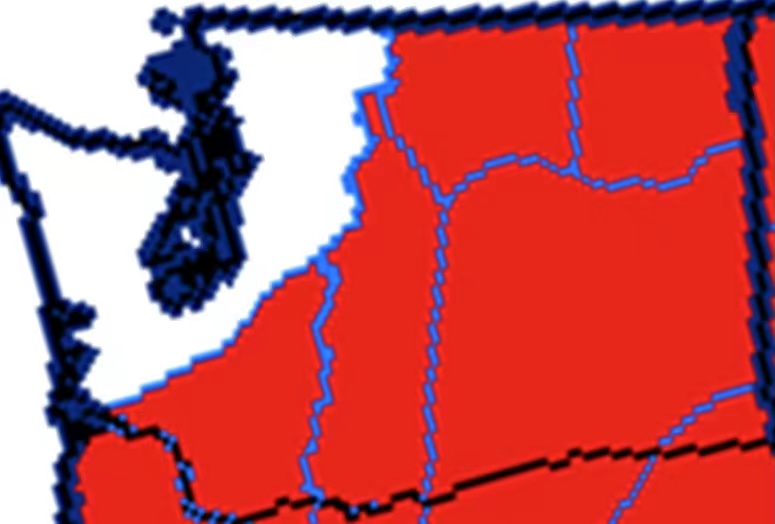}
         \caption{National Significant Wildland Fire Potential Outlook.  Red indicates the places with above normal wildfire risks predicted by the National Interagency Fire Center.}
         \label{fig:b}
     \end{subfigure}
     \hfill
     \begin{subfigure}[b]{0.3\textwidth}
         \centering
         \includegraphics[width=\textwidth]{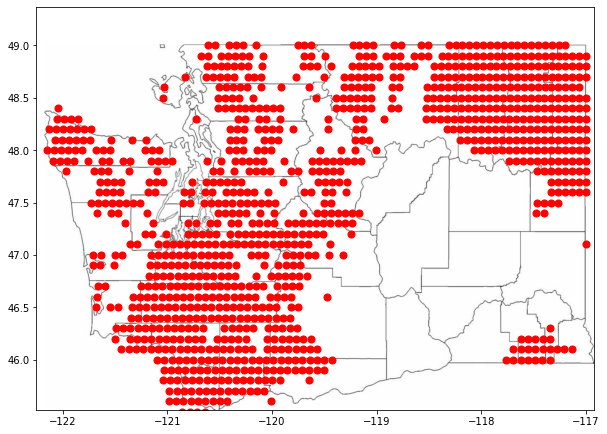}
         \caption{Our predicted high risk zones (in red).}
         \label{fig:c}
     \end{subfigure}
        \caption{Comparisons of prediction results.}
        \label{fig: washington}
\end{figure}
The confusion matrix of the prediction is presented in Table~\ref{tab:cm_evolving}. The overall accuracy is 80.60\%. The recall is 58.50\% indicating a good prediction performance in detecting grids where there is at least one wildfire incident that spreads more than 300 acres in the year of prediction. The predicted grids are plotted in the left column of Figure~\ref{fig:pre_act_ev}. Take 2013 and 2015 as examples, compared to the prediction of 2013, there are more predicted incoming wildfire risk grids in the north-west of California and much fewer around Los Angeles in 2015. In reality, there were more fire events in the north-west and fewer events in Los Angeles in 2015 than in 2013. Even though the shape of real wildfire events was not as widely spread as predicted, our trained model can predict the general incoming wildfire risk areas, and hopefully allow a prompt response to the danger.

\subsection{Prediction in 2021}
\subsubsection{Background}
To verify the generalizability of our dynamic prediction model, we repeat the same procedure using the latest data from Washington State, which is another west-coast state under a high threat of vast wildfires in recent years~\cite{everett1999snag}. On July 14, 2021, the Department of Ecology in Washington State declares drought emergency as the state encountered the second driest period since 1895~\cite{washington}. It is important to know the impact of the historical drought on the wildfire risk throughout the state.

\subsubsection{Experiment}
We collect data from the same sources as mentioned before in Washington State in mid-July 2021. {\it Without additional training}, we make the prediction using the same model as mentioned in the previous section. The latest Palmer Drought Severity (PDSI) data is posted graphically from drought.gov~\cite{drought}. Therefore, we manually extract the PDSI data from the graph posted in mid-July.

\subsubsection{Results}
As shown in Figure~\ref{fig:c}, our model marked over 1,031 out of 2,290 grids as high risk zones. The risk areas are mostly forest zones from the mid to the upper northwest of the state. On August 1, 2021, the National Interagency Fire Center published their significant wildfire potential outlook for the upcoming month~\cite{forcast} (Figure~\ref{fig:b}), and marked the entire eastern Washington State as high risk zones. By comparing to the actual Northwest Large Fire Interactive Map in August~\cite{NWCC} (Figure~\ref{fig:a}), our prediction using the multilayer neural networks manages to cover all of the actual fire areas with much fewer false positives than the official outlook.


\begin{table}[htbp!]
\centering
\caption{Confusion matrix of dynamic wildfire risk prediction.}
\begin{tabular}{|c|c|c|} \hline
 &Large Fire& No large Fire\\ \hline
Predicted High risk Zone & 137 & 3177 \\ \hline
Predicted Low risk zone & 97 & 13557 \\ \hline
\end{tabular}
\label{tab:cm_evolving}
\end{table}

\begin{figure}
  \includegraphics[width=\linewidth]{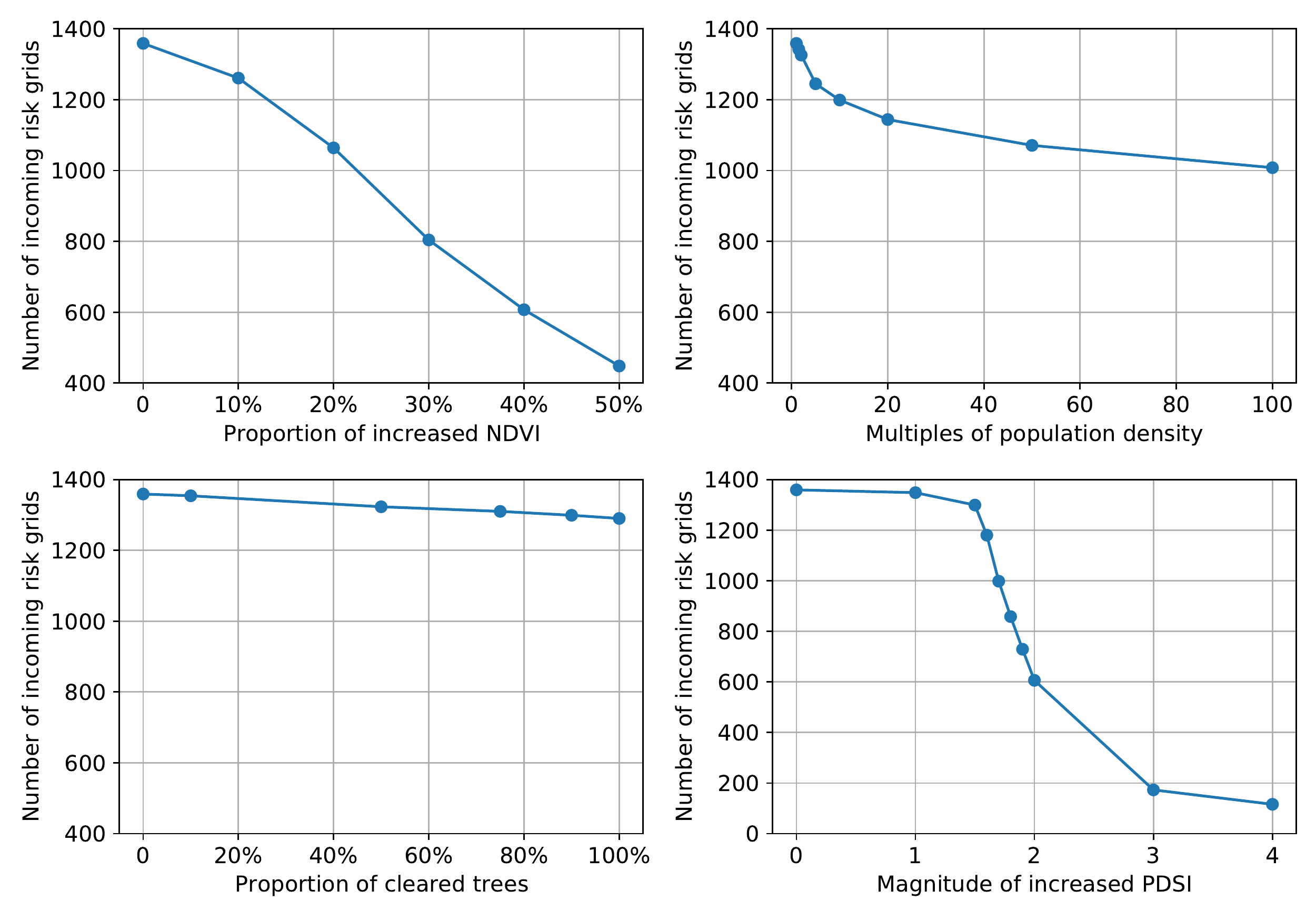}
  \caption{The impact of changing different factors.}
  \label{fig: comparison}
\end{figure}


\begin{figure*}[htbp!]
    \centering
    \includegraphics[width = 0.65 \linewidth]{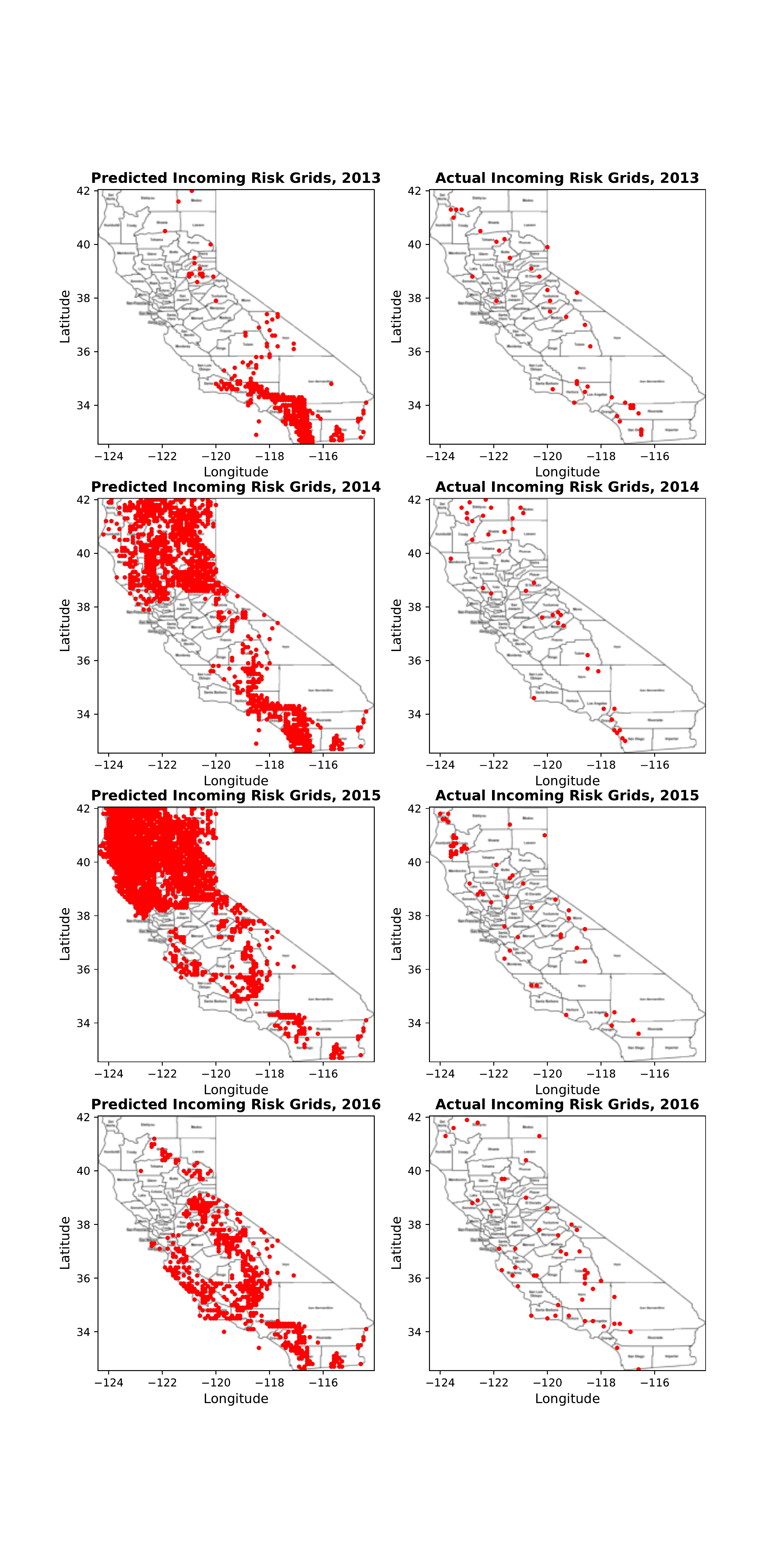}
    \vspace*{-2.5cm}
    \caption{Predicted (left) and actual (right) incoming wildfire risk grids of 2013 - 2016.}
    \label{fig:pre_act_ev}
\end{figure*}





\subsection{Counterfactual Analysis}
In this section, we attempt to study the effects of changing each of the factors (i.e., NDVI, PDSI, tree mortality, and population density) that has association with the wildfire incidents. Although we can not study the causality effects directly through experiments, the prediction model allows us to investigate the changes in risk of wildfires after applying the hypothetical treatment. More specifically, we use the trained model of the dynamic wildfire risk prediction. The counterfactual analysis has been applied in many areas such as public opinion inference~\cite{lyu2021social} and mobility network~\cite{chang2021mobility}. In each scenario of our study, we change the value of one feature and keep the others constant. Next, we compare the simulated number of incoming risk grids. We take the year of 2015 as an example in which our prediction model assigns 1,359 as incoming risk grids. The complete results of the counterfactual analysis can be found in Figure~\ref{fig: comparison}.

\subsubsection{Drought control}
Although we can not control the climate, there are still many actions we could take in the high risk zones including proper irrigation that could not only relieve the drought but also increase the biodiversity in the local area~\cite{irrigation}. 
When we increase the drought severity index (reduce the drought severity) by 1.5 unit, the number of incoming risk grids is reduced by 60.

As we continue to increase the drought severity by 2, 3 and 4, the number of incoming risk grids gradually decreases to only 116 in southern California. The curve is relatively flat when we increase the PDSI by 1 unit. The number of incoming risk grids drops quickly when we increase the PDSI by 1 to 3 units. Finally, the curve becomes flat again. In particular, increasing the PDSI by 2 units reduces more than half of the high risk wildfire incidents. Conclusively, more than $80\%$ of the wildfire incidents that spread more than 500 acres can be eliminated if we reduce the drought severity in the predicted incoming wildfire risk grids in 2015, which points us to a clear direction on how to reduce the wildfire problems in California. 
\subsubsection{``Rake the forest''}
When being asked about the wildfires on the west coast, former president Donald Trump always referred to ``raking the forest'' as the ultimate solution to the problem~\cite{cbs,abc}. In general, his proposal was to remove the dead woods in the forest so that the amount of burning fuels will be reduced. With the tree mortality data, we simulate the effects of clearing the dead trees in the past two years and analyze if it is an effective strategy to reduce the number of wildfire incidents. As shown in Figure~\ref{fig: comparison}, even after removing all the deceased trees in 2015, the number of incoming risk zones only reduce by 36 from 1,359 to 1,323. 

\subsubsection{``Composition of healthy trees''}
According to the Earth Observing System (EOS~\cite{EOS}), NDVI is not only a good indicator on land cover, but also a good indicator of trees' health. NDVI is calculated by the equation $ \frac{NIR-RED}{NIR+RED} $ where NIR stands for reflection in the near-infrared spectrum, and RED stands for reflection in the red range of the spectrum. Compared to the healthy trees, unhealthy trees will reflect $10\%$ less NIR while reflecting up to $22\%$ more RED, both of which contribute to a decline in the NDVI~\cite{EOS}. In our analysis, we investigate the effect of more healthier trees in the high risk zones by increasing the NDVI by $10\%, 20\%, 30\%, 40\%$ and $50\%$. In the treatment groups, the number of incoming risk grids shrinks from 1,359 gradually to 448. The rate of descent is relatively constant compared to the sudden and rapid decline when we increase the annual PDSI. 
\subsubsection{Population density management}
Previous studies show that the wildfire risk reaches the maximum when the human activity level is intermediate~\cite{human}. Surprisingly, in our prediction model, reducing the population density (even down to 0) does not have an impact on reducing the risk. This can be partially explained by the fact that many wildfires indeed occur deep in forests where the population is close to 0, so the prediction model would perceive the decline in population density as an increase in risk. However, increasing the population density dramatically (e.g. by 10 times) has a moderate effect that reduces the number of incoming risk girds from 1,359 to 1,199. The results show that it would be beneficial to reduce the risk of wildfires if most people are aggregating in small communities rather than scattering in different places. 

\section{Conclusions}
In this work, we integrate data from five different sources into static and dynamic prediction models to analyze and assess wildfire risks. Using Multi-layer Neural Network, Support Vector Machine, Logistic Regression, and Random Forest, we have successfully classified the geography of California into high, medium and low risk zones using a multitude of environmental predictors including PDSI, altitude, NDVI, population density and tree mortality based on all historical data. After adding the time dimension, we train another Multi-layer Neural Network to predict the wildfire risks of each year based on the same predictors that are specified by the year. The generalizability of our models is further validated by applying them, without any further tuning, to assessing the wildfire risks in Washington State. In the end, we use the trained model of the dynamic wildfire risk prediction to reveal the effect of each factor, and make practical suggestions on effective ways to reduce wildfire risks in the future.

%
\bibliography{main}

\end{document}